\documentclass[letterpaper]{article} 
\usepackage{aaai2026}  
\usepackage{times}  
\usepackage{helvet}  
\usepackage{courier}  
\usepackage[hyphens]{url}  
\usepackage{graphicx} 
\usepackage{natbib}  
\usepackage{caption} 
\usepackage{algorithm}
\usepackage{algorithmic}
\usepackage{multirow}
\usepackage[table,xcdraw]{xcolor}
\usepackage{colortbl}
\usepackage[graphicx]{realboxes}
\usepackage[figuresleft]{rotating}
\usepackage{newfloat}
\usepackage{listings}
\DeclareCaptionStyle{ruled}{labelfont=normalfont,labelsep=colon,strut=off} 
\floatstyle{ruled}
\newfloat{listing}{tb}{lst}{}
\floatname{listing}{Listing}
\title{A Benchmark Dataset for Spatially Aligned Road Damage Assessment in Small Uncrewed Aerial Systems Disaster Imagery}
\author{
    Thomas Manzini\equalcontrib, Priyankari Perali\equalcontrib, Raisa Karnik, Robin R. Murphy
}
\affiliations{
    Department of Computer Science, Texas A\&M University\\
    435 Nagle St, College Station, TX, United States, 77843\\

    \{tmanzini, perali, raisak, robin.r.murphy\}@tamu.edu
}

\usepackage{bibentry}

\begin{document}

\maketitle

\begin{abstract}
This paper presents the largest known benchmark dataset for road damage assessment and road alignment, and provides 18 baseline models trained on the CRASAR-U-DRIODs dataset's post-disaster small uncrewed aerial systems (sUAS) imagery from 10 federally declared disasters, addressing three challenges within prior post-disaster road damage assessment datasets. 
While prior disaster road damage assessment datasets exist, there is no current state of practice, as prior public datasets have either been small-scale or reliant on low-resolution imagery insufficient for detecting phenomena of interest to emergency managers. 
Further, while machine learning (ML) systems have been developed for this task previously, none are known to have been operationally validated. 
These limitations are overcome in this work through the labeling of 657.25km of roads according to a 10-class labeling schema, followed by training and deploying ML models during the operational response to Hurricanes Debby and Helene in 2024.
Motivated by observed road line misalignment in practice, 9,184 road line adjustments were provided for spatial alignment of a priori road lines, as it was found that when the 18 baseline models are deployed against real-world misaligned road lines, model performance degraded on average by 5.596\% Macro IoU.
If spatial alignment is not considered, approximately 8\% (11km) of adverse conditions on road lines will be labeled incorrectly, with approximately 9\% (59km) of road lines misaligned off the actual road.
These dynamics are gaps that should be addressed by the ML, CV, and robotics communities to enable more effective and informed decision-making during disasters.

\end{abstract}

%
\begin{links}
    \small\link{Code}{www.github.com/CRASAR/CRASAR-U-DROIDs-RDA}
    \small\link{Data \& Models}{https://huggingface.co/CRASAR}
\end{links}

\section{Introduction}
\label{sec:intro}
\begin{figure}
    \centering
    \includegraphics[width=\linewidth]{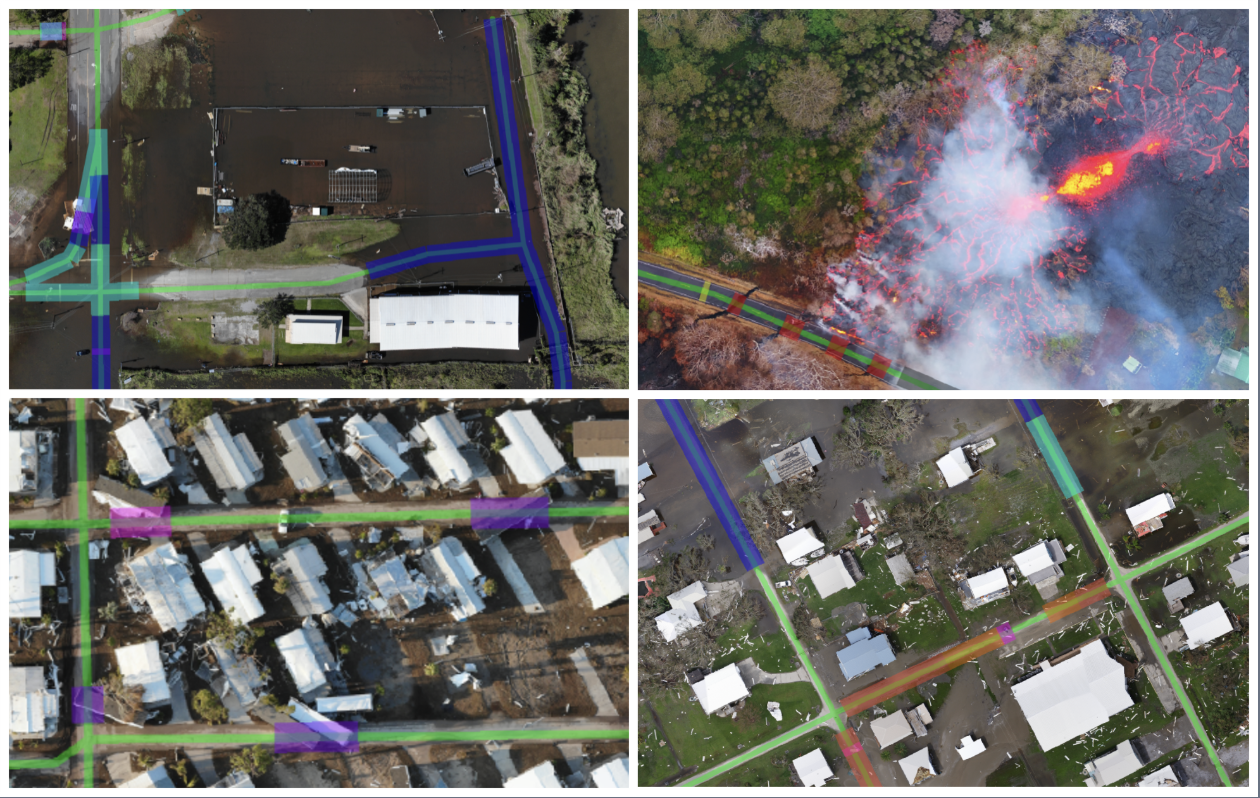}
    \caption{Figure of labeled roads within the CRASAR-U-DRIODs dataset. Road lines (Green) labeled as follows: Total Flooding (Dark Blue), Partial Flooding (Light blue), Total Obstruction (Purple), Partial Obstruction (Pink), Total Destruction (Red), Partial Road Condition (Yellow), Total Particulate (Dark Orange), Partial Particulate (Light Orange), and Not Able to Determine (Black).}
    \label{fig:introduction_fig}
\end{figure}

In response to a disaster, small uncrewed aerial systems (sUAS), also known as drones, are deployed to capture aerial imagery mapping impacted areas. This imagery is then assessed to inform emergency managers of where damage is. This information collected during the early stages of the response phase can enable informed decisions, such as appropriate allocation of aid, and effective navigation through impacted areas \cite{hargis2024search, alam2025harnessing}. However, efforts are often hindered by resource constraints and wireless connectivity \cite{manzini2023quantitative, manzini2023wireless}, arguing for Machine Learning (ML) and Computer Vision (CV) to automate assessments on hardware deployed within the disaster scene. 


This work advances the state of the art in CV/ML for sUAS post-disaster imagery by creating a labeled and spatially aligned dataset and baseline models of damaged roads, and operationally validates it with imagery collected by responders at Hurricanes Helene and Debby. Assessing road conditions is important to disaster response efforts because it enables informed decision-making concerning resource allocation, routing, and navigation of aid and evacuation. Despite its importance, prior CV/ML literature has overlooked automating post-disaster road condition assessment. 
Due to this, the field of CV/ML has not overcome three challenges in automating road damage assessments: limited diverse datasets (\cite{rahnemoonfar2021floodnet, rahnemoonfar2023rescuenet, jiang2024earthquakenet, Pi2020, hansch2022spacenet}), no operational road damage assessment schema, and the lack of operational validation of road damage assessment models. 
Furthermore, this work reveals that road damage assessment via sUAS imagery is affected by non-uniform spatial misalignment of a priori road lines, a fundamental problem prevalent within sUAS imagery collected operationally \cite{manzini2024non, manzini2025challenges}. Such non-uniform misalignment is not seen in satellite imagery, and an evaluation of 18 baseline models reveals that if spatial alignment is not addressed, model performance degrades by 1.9 Macro IoU for the top model.

This work offers five contributions to the ML, CV, remote sensing, and emergency management communities.
\begin{enumerate}
    \item The release of the largest known dataset, in terms of kilometers of roads, for road damage assessment in sUAS imagery, providing 657.25km of labeled roads.
    \item The release of the first dataset addressing spatial alignment errors with a priori road lines, providing 9,184 adjustment annotations for spatial alignment.
    \item The development of a practitioner-relevant schema for road conditions in disaster imagery, with consultation from federal and local agencies in the United States.
    \item The release of baseline models that can be used within disaster response, evaluating and validating one baseline model operationally with imagery collected in response to Hurricanes Debby and Helene.
    \item The identification of two critical challenges for future ML and CV efforts for road damage assessment concerning spatial alignment and the distribution shifts present in real-world road damage assessment aerial imagery data.
\end{enumerate}



\begin{table}
\begin{tabular}{|l|l|}
\hline
\rowcolor[HTML]{C0C0C0} 
\textbf{\small Damage Label}                          & \textbf{\small Damage Label Description}                                                                                                            \\ \hline
\cellcolor[HTML]{B8D8A9}\scriptsize Clear Road             & \scriptsize\begin{tabular}[c]{@{}l@{}}Obviously clear, road lines clearly visible, and cars \\ may be actively driving on the road surface\end{tabular} \\ \hline
\cellcolor[HTML]{F6B36C}\scriptsize Partial Obstruction    & \scriptsize\begin{tabular}[c]{@{}l@{}}Road partially covered by scattered debris, \\obstructing vegetation, piled cars, RVs, and boats\end{tabular}     \\ \hline
\cellcolor[HTML]{F6B36C}\scriptsize Partial Flooding       & \scriptsize Standing water on road, road lines partially visible                                                                                         \\ \hline
\cellcolor[HTML]{F6B36C}\scriptsize Partial Road Condition & \scriptsize Road partially crumbling, moderate asphalt cracking                                                                                          \\ \hline
\cellcolor[HTML]{F6B36C}\scriptsize Partial Particulate    & \scriptsize\begin{tabular}[c]{@{}l@{}}Road partially covered by particulates (e.g. sand,\\mud, lahar), road lines partially visible\end{tabular}       \\ \hline
\cellcolor[HTML]{EC9A9A}\scriptsize Total Obstruction      & \scriptsize\begin{tabular}[c]{@{}l@{}}Total coverage of the road by piles of debris,\\ vegetation, or vehicles preventing transit\end{tabular}         \\ \hline
\cellcolor[HTML]{EC9A9A}\scriptsize Total Flooding         & \scriptsize \begin{tabular}[c]{@{}l@{}}Large quantities of standing water on the road;\\road lines are not visible\end{tabular}                                                                    \\ \hline
\cellcolor[HTML]{EC9A9A}\scriptsize Total Destruction      & \scriptsize Road absent or collapsed, substantial asphalt cracking                                                                                       \\ \hline
\cellcolor[HTML]{EC9A9A}\scriptsize Total Particulate      & \scriptsize\begin{tabular}[c]{@{}l@{}}Total coverage of the road by particulates (e.g. sand, \\mud, lahar); road lines are not visible\end{tabular}    \\ \hline
\cellcolor[HTML]{6D9FEC}\scriptsize Not Able To Determine  & \scriptsize Unable to determine due to obscuration                                                                                                       \\ \hline
\end{tabular}
    \caption{A simplified version of the 10 classes in the Road Damage Assessment Schema. Green corresponds to the ``Road Line" class, orange corresponds to all the ``Partial" classes, red corresponds to the ``Total" classes, and blue corresponds to the ``Not Able To Determine" class.\label{tab:RDA_Schema_Table}}
\end{table}

\section{Related Work}
The remote sensing, civil engineering, ML, and CV communities have considered and developed datasets for road damage assessment and road condition schemas. However, these datasets are limited to specific road conditions, specific disasters, small data scales, and do not consider spatial misalignment errors with a priori road lines. Further, these datasets all use schemas that do not align with or capture the necessary road damage labels for disaster response, hindering the transfer of trained ML models to practice. 

\begin{figure*}
    \centering
    \includegraphics[width=0.79\linewidth]{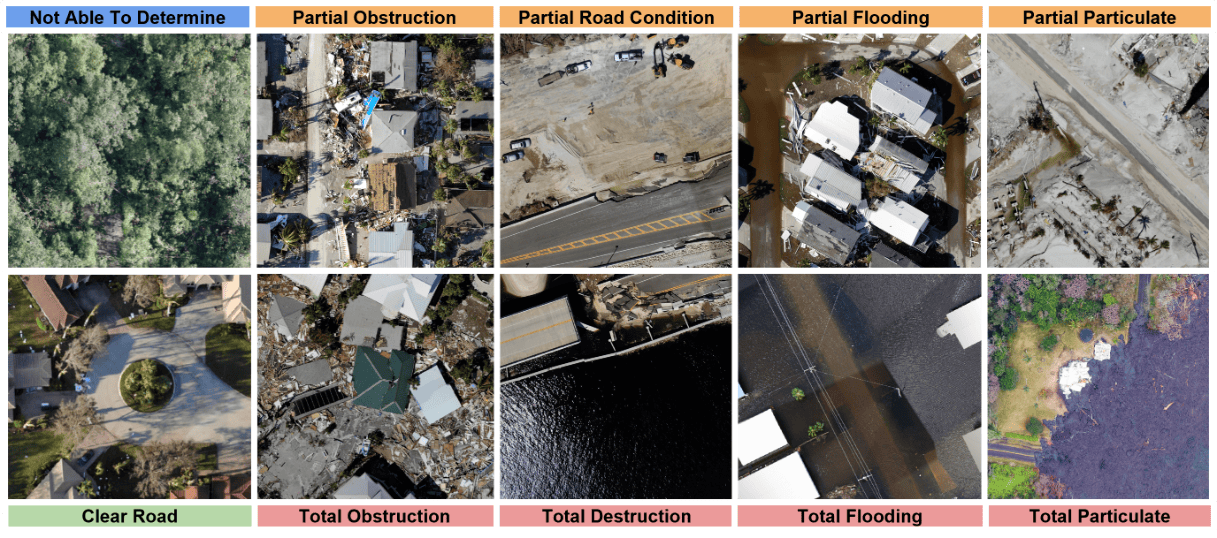}
    \caption{Visuals of the 10 Labels within the Road Damage Assessment Schema.}
    \label{fig:RDA_schema}
\end{figure*}

\subsection{Disaster Datasets for Road Damage Assessment}
Prior literature contains eight known efforts to automatically assess road conditions using aerial imagery. These consist of three datasets of sUAS aerial imagery \cite{rahnemoonfar2021floodnet, rahnemoonfar2023rescuenet, jiang2024earthquakenet}, one dataset containing satellite aerial imagery \cite{hansch2022spacenet}, and one dataset with both sUAS and crewed aircraft imagery \cite{Pi2020} with labels for flooded areas corresponding to roads. The remaining four efforts did not release data \cite{urabe2007analysis, korkmaz2016path, yang2020extraction, takyi2025towards} but considered sUAS and satellite imagery. Unfortunately, all lack practitioner-relevant labels, limiting use in disaster response.

The three sUAS-centric datasets, FloodNet \cite{rahnemoonfar2021floodnet} RescueNet \cite{rahnemoonfar2023rescuenet}, and EarthquakeNet \cite{jiang2024earthquakenet}, are semantic segmentation datasets that provide road damage labels for whether the roads were: ``flooded" or ``non-flooded", ``road clear" or ``road blocked", and ``no damage", ``minor damage", or ``severe damage", respectively. FloodNet provides labels for 31.58km of annotated road, consisting of 2.85km\textsuperscript{2} and 28.12 gigapixels of imagery from Hurricane Harvey. RescueNet provides labels for 41.05km of road, consisting of 3.6km\textsuperscript{2} and 53.93 gigapixels of imagery from Hurricane Michael. EarthquakeNet provides labels for approximately 1.4 gigapixels of imagery of roads from the 2013 Lushan Earthquake in Sichuan, China.

The remaining five efforts use similar schemas, ``flooded" and ``non-flooded" \cite{hansch2022spacenet}, ``road blockage" and ``no road blockage" \cite{yang2020extraction, urabe2007analysis}, ``open," ``partially-open,"  and ``undamaged," ``slightly damaged," and ``damaged"  \cite{korkmaz2016path}, and ``debris," ``flooded area," and ``car" \cite{Pi2020} for specific disasters and road conditions. The most diverse dataset among these is the Volan v.2018, with three classes and imagery from hurricanes Harvey, Irma, Maria, and Michael; it does not overcome the earlier limitations.

\subsection{Dedicated Road Condition Schemas}
The civil engineering literature has provided a standardized schema, the ``pavement condition index," for analyzing pavement condition, with four datasets \cite{majidifard2020pavement, sabouri2023machine, ren2024annotated, yan2023uav} motivated to automate such analysis. The works in this area focus primarily on measuring the health of the pavement or estimating its lifespan; this is different from this work, which focuses on assessing which roads could be utilized immediately following a disaster.

More generally, the transportation disruption ontology provided by \cite{corsar2015transport} provides a comprehensive ontology of events that could disrupt planned travel. While this ontology does have entries that could be relevant to this effort, such as ``Storm Damage," ``Tornado," and ``Forest Fire," these entries do not describe the conditions that are present on the road itself, nor do they describe visual features, only transportation disruptions, limiting the translation of such a schema to aerial imagery.

\section{Method}
\label{sec:method}
The method to address the three challenges, described earlier, consists of the development of a practitioner-relevant road damage schema for disaster aerial imagery, annotation of the CRASAR-U-DRIODs \cite{manzini2024crasar} imagery for road damage assessment, adjustment annotation for spatial alignment errors with the use of a priori road lines, and development of baseline models trained on this dataset. 

\subsection{Road Damage Assessment}
The road damage labels within this dataset were based on a schema developed with input from federal and state agencies in the United States, applied to imagery from the CRASAR-U-DRIODs dataset by a pool of 130 annotators, and reviewed through a two-stage process to reduce label noise. 

\subsubsection{Imagery}
The CRASAR-U-DROIDs was selected as the source of all imagery for this effort \cite{manzini2024crasar}. The motivation for this was twofold. First, this dataset is the largest known collection of sUAS orthomosaic imagery, comprising 52 orthomosaics from 10 different federally declared disasters, collected with resolutions ranging from 12.7cm/px to 1.93cm/px. Ideally, this diverse data would enable performant and generalizable ML models.
Second, this dataset contained a substantial variety of road conditions representative of real-world road conditions.

\subsubsection{Schema Formulation}

The road damage assessment schema was developed with input from the United States Federal Emergency Management Agency (FEMA) and the Texas Department of Transportation and consists of 10 damage classes, visually shown in Figure \ref{fig:RDA_schema}. The purpose of this collaboration was to ensure that models derived from this data could provide practitioner-relevant labels.

This schema contains 10 classes, eight of which are split into two subcategories: ``Partial" and ``Total". The labels of ``Road Line" and ``Not Able To Determine" are individual labels without subcategories. Otherwise, roads can be assigned to the ``Partial" class for roads that are only partially affected by an adverse condition or the ``Total" class for roads that are completely affected by an adverse condition. There are four categories of adverse conditions, which can be either ``Partial" or ``Total," and combined, totaling the eight classes referenced above. These are ``obstruction," ``flooding," ``particulate," and ``road condition/destruction." 

Obstruction describes discrete physical objects on the road's surface. Flooding describes water on the road's surface. Particulate denotes small particles that could potentially obscure conditions below them. Examples of particulates include beach sand, mud, or dirt. Lahar, while not a particulate, was intentionally folded into this class. Finally, road condition/destruction describes a condition of the road surface itself, and is either ``Road Condition" or ``Destruction" depending on the ``Partial" or ``Total" subclass. With ``Partial Road Condition," the road may still be usable but has cracked or is crumbling away, whereas in ``Total Destruction," the road has been destroyed and cannot be used.

\subsubsection{Annotation}
\begin{figure*}
    \centering
    \includegraphics[width=0.8\textwidth]{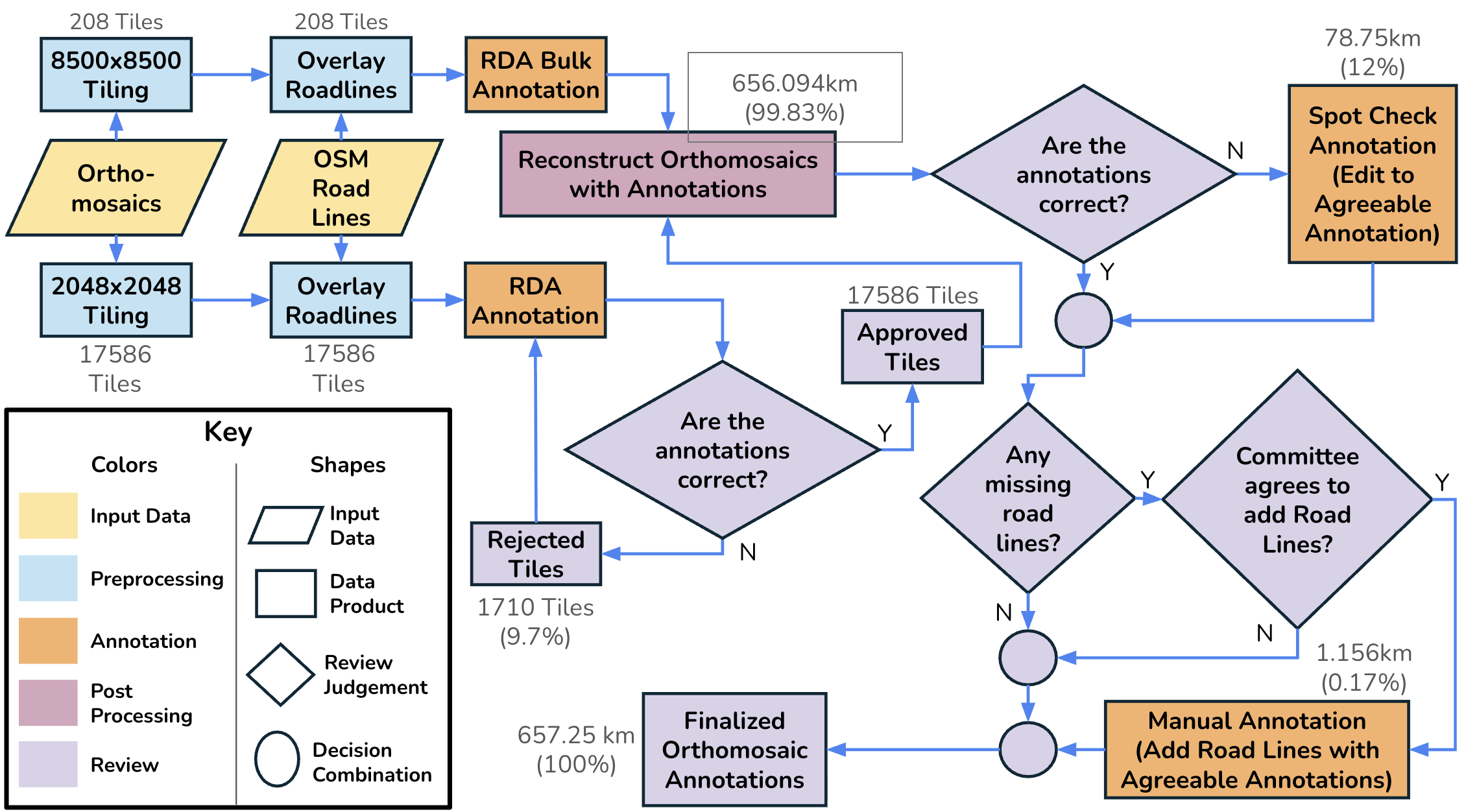}
    \caption{The visualized annotation workflow by which unlabeled imagery is labeled, reviewed, and processed.}{\label{fig:RDA_annotation_workflow}}
\end{figure*}

The imagery within the CRASAR-U-DRIODs dataset was tiled, overlaid with a priori road lines, sourced from OpenStreetMap (OSM) \cite{OpenStreetMap}, and annotated by a pool of 130 annotators, within Labelbox\cite{Labelbox2024}, according to the schema discussed earlier, resulting in 656.094km of road annotated. Each of the 52 orthomosaics within the dataset was tiled at a resolution of 2048x2048 (45 orthomosaics) and 8500x8500 (7 orthomosaics). While the pool of 130 annotators annotated the 2048x2048 tiles, the 8500x8500 tiles were annotated by the authors. A view of this process is shown in Figure \ref{fig:RDA_annotation_workflow}.
Though the annotators were non-experts, reviewers compensated for the lack of domain expertise.

\subsubsection{Annotation to Road Corridor Processing}
The road condition annotation polygons provided by the annotators were often of inconsistent shape and, in some cases, did not correspond to any road lines. To make these annotation polygons more consistent and to align them with the road itself, the intersection between all road lines and all annotation polygons was computed. Then, new rectangular polygons were generated based on these intersections such that they were parallel with the road line and had a width of 7.2 meters, the reported upper limit on the width of two lanes of rural or urban road in the United States \cite{mitigation2007}. Examples of these rectangular polygons are shown in Figure \ref{fig:introduction_fig}. This intentional decision was to first standardize the dimensions of the annotation polygon, so they corresponded to the dimensions of roads rather than the dimensions that annotators had drawn, and second, to have a dimension that could provide a consistent region to account for reasonable variations in spatial misalignment, discussed later on. 

\subsubsection{Review}
\begin{figure}
    \centering
    \includegraphics[width=\columnwidth]{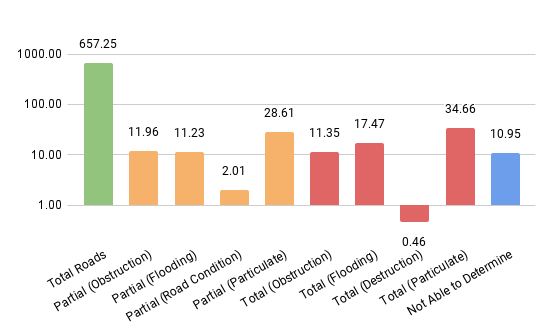}
    \caption{Kilometers of each label in the presented dataset. Note that the y-axis is on a log scale.}
    \label{fig:RDA_stats}
\end{figure}
\label{sec:reviews}
Following annotation, a two-stage review was conducted: first, a review by individual reviewers, and second, an orthomosaic review by a committee of reviewers. During the individual review, each tile was inspected by a single reviewer, and 1,710 tiles (9.7\% of tiles) were rejected and re-annotated by annotators or corrected by the reviewer. Following this, the annotations were overlaid on orthomosaics and reviewed by a committee consisting of three of the authors and one external reviewer, where spot-check corrections were made. Reviewers collectively had experience working operationally at major disasters, working as an emergency vehicle operator (EVOC/CEVO), a sUAS data manager, and a sUAS pilot. Additionally, at the discretion of the reviewers, missing road lines were manually added and annotated. The spot-check corrections resulted in 78.5km (12\%) of road line labels being changed. The manual addition of road lines, not sourced from OSM, resulted in 1.156km (0.17\%) of road lines being added. Following this review, the distribution of labels shown in Figure \ref{fig:RDA_stats} remained. This two-stage review process was employed to maximize consistency and minimize label noise, with the goal of creating a worthwhile benchmark for future models. An overview of this process can be found in Figure \ref{fig:RDA_annotation_workflow}.

\subsection{Spatial Alignment}
\begin{figure}
    \centering
    \includegraphics[width=\columnwidth]{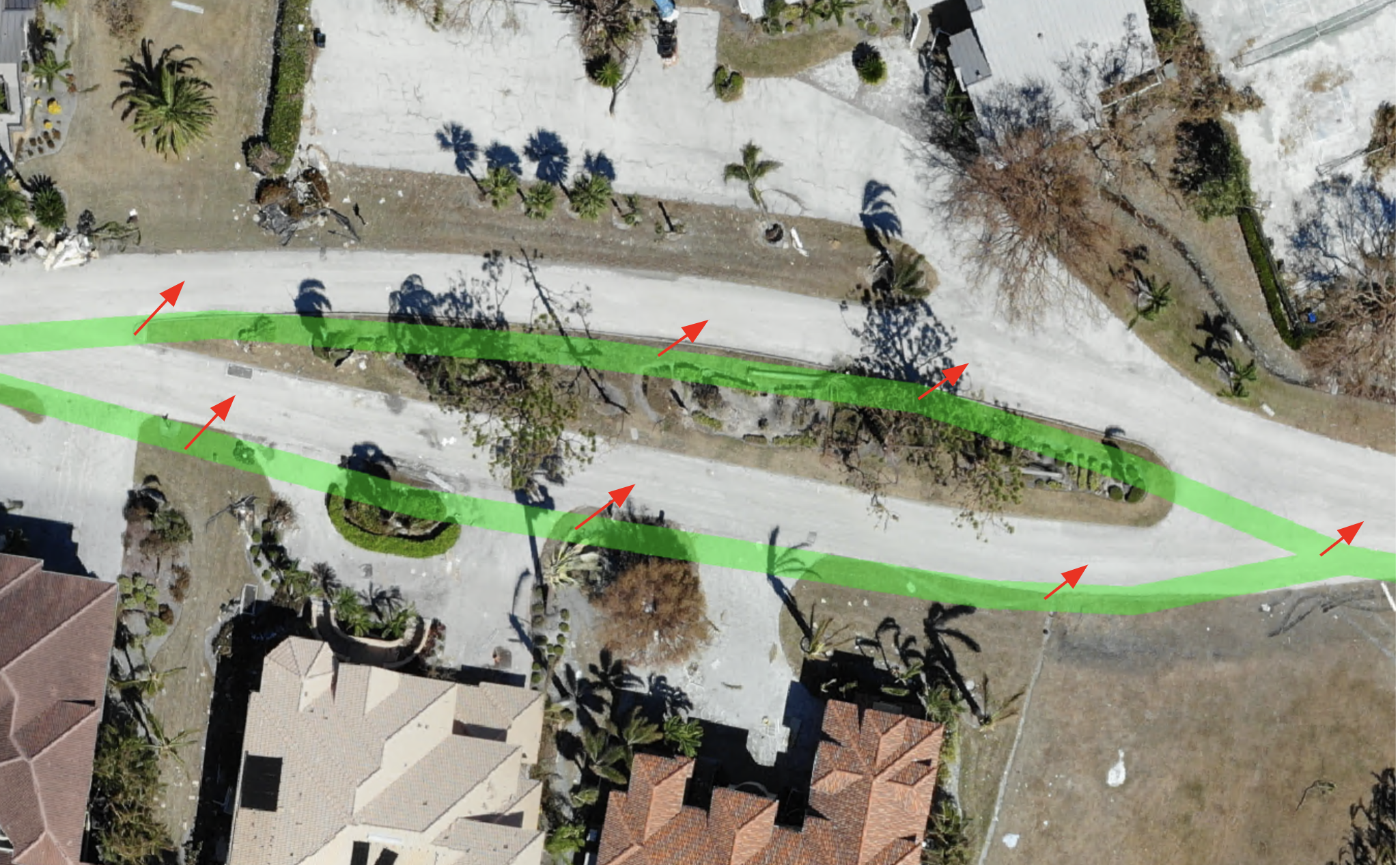}
    \caption{Example of a road line (colored in green) that is unaligned with the source imagery. Red alignment vectors show the transform necessary to align each vertex.}
    \label{fig:alignment_example}
\end{figure}

During the curation of the road damage assessment labels, spatial alignment errors with the a priori road lines were observed, an example is shown in Figure \ref{fig:alignment_example}. Approximately 59km (9\%) of road lines were misaligned off the actual road, and 11km (8\%) had incorrect road damage assessment labels, presenting the need to correct the errors to avoid impeding downstream ML models trained on this data, resulting in 9,184 road line adjustment annotations. 

This work aligns road lines using the same vector field formulation presented in \cite{manzini2024non}, but differs by operating on individual road line vertices instead of building polygons. While the CRASAR-U-DRIODs dataset provides adjustment annotations for similar misalignment with a priori building polygons, these were generated for the building polygons, which were sourced from a different spatial resource than the road lines. As a result, additional adjustments were manually collected and curated for the road lines considered in this work, resulting in 9,184 total road line adjustments.
As alignment of road lines acts on vertices, the aligned road lines differ from the misaligned by 396 meters across all 657.25km of road lines (0.06\%).

\subsection{Annotation Format}
The annotations are presented in the form of road lines and annotation polygons. This is an intentional choice to enable these annotations to be robust to future efforts to model and thereby vary spatial alignment. In this format, road lines can be annotated by computing the intersection of road lines with the annotation polygons. This decouples road line annotations from variations in road line alignment, meaning that ML models predicting road line labels can generate ground truth labels independently of alignment logic.

\subsection{Baseline Models}
\label{sec:baselines}
This work introduces eighteen baseline models, each of which formulates road damage assessment as a segmentation task, as benchmarks for future investigations. These eighteen are composed of nine baseline model architectures, each trained and evaluated on the ``Simple" and ``Full" prediction tasks, defined later. These nine architectures are as follows: 1) A Random Baseline, 2) UNet without attention  \cite{ronneberger2015u}, 3) UNet with attention \cite{oktay2018attention}, 4) ResNet101 \cite{he2016deep}+ PSPNet \cite{zhao2017pyramid}, 5) ResNet101 \cite{he2016deep} + DeepLabv3plus \cite{chen2018encoder}, 6) Vision Transformer \cite{reed2023scale} + Segmenter \cite{strudel2021segmenter}, 7) Vision Transformer \cite{reed2023scale} + Segmenter (Pretrained Vision Transformer)\cite{strudel2021segmenter}, 8) Vision Transformer \cite{reed2023scale} + UperNet \cite{xiao2018unified}, 9) Vision Transformer \cite{reed2023scale} + UperNet (Pretrained Vision Transformer) \cite{xiao2018unified}. 

The sixteen trainable baselines provide guidepost performance for future efforts, and the random baseline provides a reasonable lower bound. Road lines were masked by buffering all road lines by 40 pixels, forming a rectangular mask, and all trained models utilized two strategies simultaneously for class imbalance mitigation: weighted sample presentation targeting uniform label propensity and CCE loss weighted by observed inverse label propensity.




\begin{table*}[]
\centering
\resizebox{\textwidth}{!}{
\begin{tabular}{|l|rrrr|rrrr|}
\hline
\multirow{3}{*}{\textbf{Model}}                                                                                                                     & \multicolumn{4}{c|}{\textbf{Simple}}                                                                                                                                    & \multicolumn{4}{c|}{\textbf{Full}}                                                                                                                                      \\ \cline{2-9} 
                                                                                                                                                    & \multicolumn{2}{c|}{\textbf{Adjusted}}                                              & \multicolumn{2}{c|}{\textbf{Unadjusted}}                                          & \multicolumn{2}{c|}{\textbf{Adjusted}}                                              & \multicolumn{2}{c|}{\textbf{Unadjusted}}                                          \\
                                                                                                                                                    & \multicolumn{1}{c}{\textbf{IoU$_{km}$}} & \multicolumn{1}{c|}{\textbf{F1$_{km}$}}   & \multicolumn{1}{c}{\textbf{IoU$_{km}$}} & \multicolumn{1}{c|}{\textbf{F1$_{km}$}} & \multicolumn{1}{c}{\textbf{IoU$_{km}$}} & \multicolumn{1}{c|}{\textbf{F1$_{km}$}}   & \multicolumn{1}{c}{\textbf{IoU$_{km}$}} & \multicolumn{1}{c|}{\textbf{F1$_{km}$}} \\ \hline
\textbf{UNet with Attention} \cite{oktay2018attention}                                                            & {\underline{\textbf{0.331}}}                    & \multicolumn{1}{r|}{{\underline{\textbf{0.393}}}} & {\underline{\textbf{0.312}}}                    & { \underline{\textbf{0.368}}}                    & \textbf{0.091}                          & \multicolumn{1}{r|}{0.095}                & \textbf{0.092}                          & 0.096                                   \\ \hline
\textbf{UNet w/out Attention} \cite{ronneberger2015u}                                                             & 0.307                                   & \multicolumn{1}{r|}{0.376}                & 0.279                                   & 0.342                                   & \textbf{0.091}                          & \multicolumn{1}{r|}{0.095}                & \textbf{0.092}                          & 0.096                                   \\ \hline
\textbf{ResNet101 + DeepLabv3Plus} {[}\cite{he2016deep} + \cite{chen2018encoder}{]}              & 0.165                                   & \multicolumn{1}{r|}{0.264}                & 0.143                                   & 0.234                                   & 0.081                                   & \multicolumn{1}{r|}{{\underline{\textbf{0.103}}}} & 0.076                                   & 0.096                                   \\ \hline
\textbf{ResNet101 + PSPNet} {[}\cite{he2016deep} + \cite{zhao2017pyramid}{]}                     & 0.283                                   & \multicolumn{1}{r|}{0.388}                & 0.269                                   & 0.370                                   & 0.084                                   & \multicolumn{1}{r|}{0.095}                & 0.083                                   & 0.095                                   \\ \hline
\textbf{ViT-L + Segmenter} {[}\cite{reed2023scale} + \cite{strudel2021segmenter}{]}              & 0.039                                   & \multicolumn{1}{r|}{0.074}                & 0.033                                   & 0.064                                   & \textbf{0.091}                          & \multicolumn{1}{r|}{0.095}                & \textbf{0.092}                          & 0.096                                   \\ \hline
\textbf{ViT-L (Pretrained) + Segmenter} {[}\cite{reed2023scale} + \cite{strudel2021segmenter}{]} & 0.015                                   & \multicolumn{1}{r|}{0.028}                & 0.013                                   & 0.025                                   & 0.091                                   & \multicolumn{1}{r|}{0.095}                & 0.091                                   & 0.095                                   \\ \hline
\textbf{ViT-L + UperNet} {[}\cite{reed2023scale} + \cite{xiao2018unified}{]}                     & 0.211                                   & \multicolumn{1}{r|}{0.309}                & 0.200                                   & 0.296                                   & 0.087                                   & \multicolumn{1}{r|}{0.099}                & 0.087                                   & \textbf{0.099}                          \\ \hline
\textbf{ViT-L (Pretrained) + UperNet} {[}\cite{reed2023scale} + \cite{xiao2018unified}{]}        & 0.209                                   & \multicolumn{1}{r|}{0.313}                & 0.185                                   & 0.284                                   & 0.087                                   & \multicolumn{1}{r|}{0.098}                & 0.087                                   & 0.098                                   \\ \hline
\textbf{Random Baseline}                                                                                                                            & 0.135                                   & \multicolumn{1}{r|}{0.215}                & 0.135                                   & 0.215                                   & 0.016                                   & \multicolumn{1}{r|}{0.031}                & 0.016                                   & 0.031                                   \\ \hline
\end{tabular}}
\caption{Model Performance for Baseline Models. Model architectures with an encoder and decoder architecture follow the naming convention ``Encoder + Decoder". The macro IoU$_{km}$ and macro F1$_{km}$ are reported for adjusted and unadjusted configurations. Bold values represent the maximum value per column. Underlined values represent metrics where any class in the macro average is significantly different from the next highest-performing model based on a Hoefding Bound with $p<0.001$.}
\label{tab:model_results}
\end{table*}

\section{Evaluations}
\label{sec:experiments}
Two evaluations were conducted to establish a range of performances that for future model development could reference and to validate model performance on real-world data. This section introduces two labeling tasks associated with this dataset. The first task, termed ``Simple," is a labeling task where the model must label road lines in one of three classes: no annotation, a ``Partial" class label, or a ``Total" class label, which also includes the ``Not able to Determine" label. In the second task, termed ``Full," the model must label the road line according to the exact ground truth label. 

The first evaluation was to establish baseline performance on this dataset using the two approaches that were described earlier, concerning the labeling tasks ``Simple" and ``Full. The second experiment was to establish the value of adjustments in this dataset. As discussed in the literature, misalignment between imagery and preexisting spatial data can result in performance degradations \cite{maiti2022effect, vargas2019correcting}. This experiment aims to determine the scale of the gap that may exist in this dataset, given the baselines. This section begins with a discussion of evaluation metrics, followed by the details and results of each of the evaluations conducted.

\subsection{Evaluation Metrics}

Two metrics are used to evaluate the performance of these ML baselines: Intersection over Union (IoU) and F1. IoU was determined to most closely align with the expected use cases of the model, while F1 is presented for additional context. In this context, IoU represents the proportion of a labeled segment of road that corresponds to the label.

Computing IoU and F1 is confounded by the variable spatial dimensions within CRASAR-U-DROIDs, with imagery varying in resolution between 12.7cm/px and 1.77cm/px. Ideally, any evaluation metric would be robust to these changes so as not to be biased by imagery near the extremes of these spatial dimensions. To combat this, evaluation takes place along this spatial dimension, rather than the pixel dimension. To compute IoU, the value of the true positive length of a road line label is its length in kilometers, rather than pixels. This decision shifts the metrics into a dimension of greater value to practitioners compared to pixels. This metric will be denoted as IoU$_{km}$ and F1$_{km}$.

\subsection{Baseline Model Performance}
\label{sec:baseline_exp}
All sixteen trainable baseline models were trained on the road line labels from the CRASAR-U-DROIDs training set orthomosaics and then tested, alongside the Random baseline, on the road line labels from the CRASAR-U-DROIDs test set orthomosaics. A complete breakdown of the macro IoU$_{km}$ and F1$_{km}$ metrics for each of the prediction tasks is shown in Table \ref{tab:model_results}, with more detailed metrics appearing in the appendix. The model performance appears to degrade as the number of prediction classes increases. 

\subsection{Evaluation of Spatial Alignment}
To measure the value that adjustments add quantitatively, the same eight baseline models trained for the baseline evaluation, described earlier, were evaluated on the test set but without any alignment. Thus, road lines would be translated out of their correct position and potentially over features in the imagery that do not correspond to the road or its conditions. The results of this evaluation are shown in Table \ref{tab:model_results}. Comparing each line between the IoU$_{km}$ adjusted and IoU$_{km}$ unadjusted, on average, the model performance degrades by 0.014 units of IoU$_{km}$ for the ``Simple" labeling task and degrades by 0.0003 units of IoU$_{km}$ for the ``Full" labeling task. As for comparing each line between the F1$_{km}$ adjusted and F1$_{km}$ unadjusted, the model performance degrades by 0.018 units of F1$_{km}$ for the ``Simple" labeling task and by 0.0004 units of F1$_{km}$ for the ``Full" labeling task.


\subsection{Evaluation at Hurricanes Debby \& Helene}
In response to Hurricanes Debby and Helene, the Attention UNet baseline, configured for the ``Simple" task, was deployed operationally in Florida to assess the effectiveness of models in practice and to gather qualitative feedback from practitioners in the field. Hurricane Debby was a Category 1 Hurricane that impacted Florida in August 2024 \cite{Debby2024}. Hurricane Helene was a Category 4 Hurricane that impacted Florida in September 2024 \cite{Helene2024}. During the response to both Hurricanes, sUAS were deployed to collect aerial imagery of the impacted areas, and the Attention UNet model was used to label imaged roads. 


\begin{figure*}
    \centering
    \includegraphics[width=0.95\linewidth]{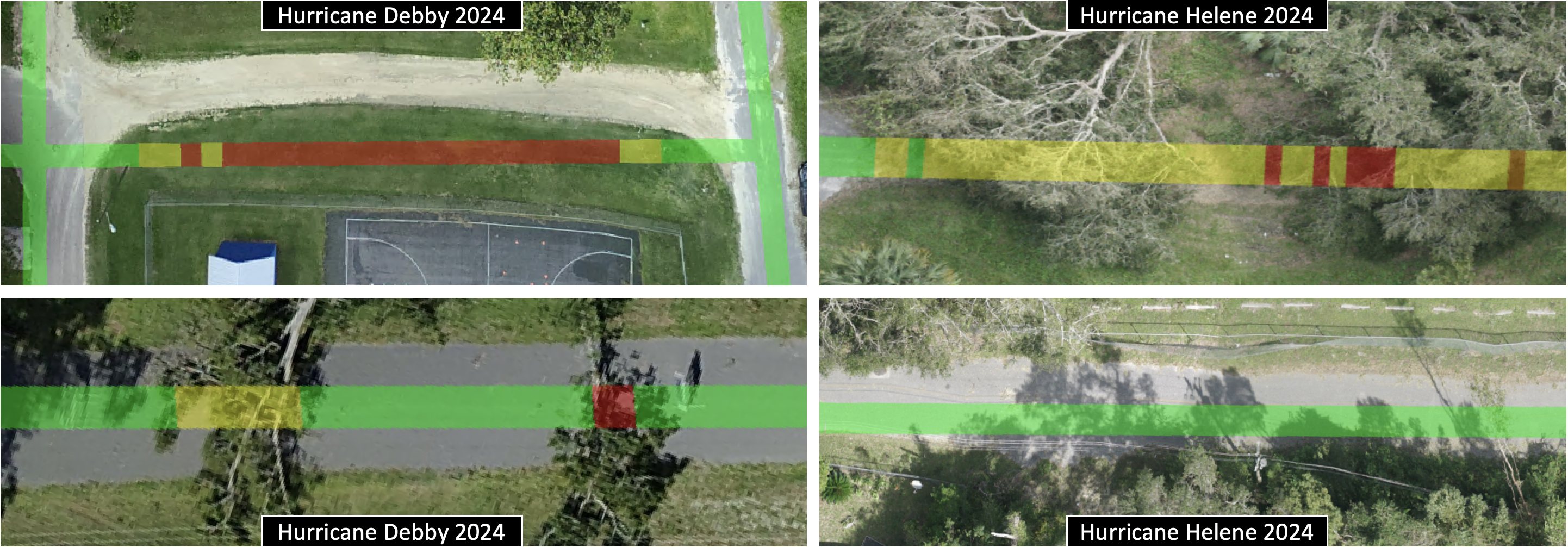}
    \caption{Sample outputs from the Attention UNet on sUAS data collected during the response to Hurricanes Debby (left) and Helene (right) in Florida. Top left: An instance where the model mislabels a road due to an alignment error. Top Right: An instance where a fallen tree blocks and obscures the road, and the model identifies obstructions. Bottom Left: An instance where the model identifies trees that have fallen onto the road. Bottom Right: a clear road that the model correctly identifies.}
    
    \label{fig:debby_sample}
\end{figure*}

Samples from the model outputs are included in Figure \ref{fig:debby_sample} as part of a qualitative assessment. This figure shows four images: two instances of fallen trees on roads that were correctly identified, one instance where misalignment results in mislabeling, and one instance where a clear road is correctly identified despite shadows and artifacts. This model processed an approximately 3mi$^2$ orthomosaic in 5 minutes on a desktop with an NVIDIA RTX4090, hardware that could reasonably be fielded alongside sUAS teams. Model outputs were converted to KML files for dissemination in practice. 

In conversations with disaster practitioners following these deployments, two important findings were made. First, false positives were far more tolerable in practice than false negatives, as practitioners immediately inspected indications of damage, but largely ignored labels for clear roads. This was because practitioners wanted to verify if labeled roads would actually be a problem for their vehicles or route, while the large prevalence of labeled clear roads led to that label being largely ignored. Second, it was found that there was no singular value for model performance that would be deemed acceptable in practice. Instead, the value of the model's predictions was time-dependent, with practitioners tolerating more errors if model outputs were delivered earlier. It was determined that the Attention UNet model for the ``Simple" task was able to provide operational value in the early stages of disasters when information is especially limited, providing strong evidence of generalization, though practitioners consistently stated they would have preferred a full schema.

\section{Discussion}
A discussion of the above evaluations must be conducted to better characterize the models' performances and to highlight spatial alignment's impact on model performance.

\subsection{Baseline Model Performance}
The low absolute performance of the trainable baseline models warrants further discussion, as it currently limits applicability. While the Attention UNet model and other baselines outperform the random baseline across every prediction task, performance remains low compared to prior work \cite{rahnemoonfar2023rescuenet, rahnemoonfar2021floodnet}. While the target function may indeed be noisy, the two-stage review process was used to mitigate this concern. Though it is reasonable to assume that some amount of label noise exists, it is believed that this contribution is minimal. Instead, the framing of this problem as a segmentation task may be a culprit, as many instances of the labels (e.g., partial obstruction, partial road condition) feature pixels that would otherwise correspond to clear roads and thus require reliance on a small number of pixels for correct classification. With high-resolution imagery, this means that models require large receptive fields to provide correct labels. As a result, it is believed that this work represents a rich space for exploring vision models' capabilities in utilizing large contexts and receptive fields.

\subsection{Relevance of Class Imbalance}
One of the challenges in this dataset is class imbalance, as shown in Figure \ref{fig:RDA_stats}. Given that CRASAR-U-DROIDs sources imagery from real-world disasters and operations, it is unsurprising that this dataset has such class imbalances. With this in mind, this dataset is representative of the types and diversity of data expected in real-world operations, as imagery is sourced directly from that distribution.


\subsection{Importance of Spatial Alignment}
Two of the evaluations conducted show a clear need for future efforts to focus on alignment. First and most clearly, the decrease shown between the baseline models when run on aligned and unaligned data is explicit quantitative evidence that model performance degrades when alignment is not managed. Second, however, is the qualitative evidence shown from the model deployment on imagery collected from Hurricane Debby. The top left image in Figure \ref{fig:debby_sample} shows unambiguously the model labeling an otherwise clear road incorrectly because the road line is not correctly aligned with the road. Additionally, an analysis of the dataset indicates that when adjustments are not applied, then 9\% of all roads (59km) fall outside of the 7.2 meter nominal width of two lanes of road \cite{mitigation2007}, and 8\% (11km) of adverse road conditions would be labeled incorrectly as they fall outside of this same 7.2 meter bound. Operationally, should ML systems be deployed without oversight, spatial misalignment may reduce route efficiency or cause spurious repair dispatches due to misalignment-driven false positives. Further, practitioners distrusted model outputs when road lines were not coincident with imagery. Therefore, ML models must manage spatial misalignment to effectively assess adverse conditions on predefined road lines in practice.

\section{Conclusion}
This paper addressed three challenges in the development of CV/ML systems for road damage assessment with post-disaster sUAS imagery, providing the infrastructure to facilitate the further development of CV/ML systems to support decision-making during disaster response. A practitioner-relevant road damage assessment schema was developed with consultation of federal and state agencies to align with operational needs. From which the largest collection of road damage assessment labels dataset was released, along with baseline models trained on this dataset. Finally, a qualitative analysis operationally validates one baseline model in response to Hurricanes Debby and Helene and finds that practitioners tolerate false positives for earlier road damage assessment, providing insight into the evaluation of such a model for operational value. 
Future work will focus on first, collecting additional data, potentially including non-disaster imagery, to improve the performance of the presented models; second, conducting analyses of the baseline models and experimenting to improve model performance and explore performance degradations on the ``Full" formulation; third, and finally, evaluating the ethical implications of both false negatives and false positives on disaster operations. This work is expected to enable the development of operationally valid CV/ML systems for disaster response, pushing the CV/ML communities to deliver systems that will support decision-making during disaster response. 


\section*{Appendix}
To further contextualize the performance of the baseline models trained within this work, additional tables are provided to inspect the performance of each model. To measure the performance of each model on each disaster Table \ref{tab:model_perf_disaster_simple} and Table \ref{tab:model_perf_disaster_full} are presented. Table \ref{tab:model_perf_disaster_simple} shows the performance of the baseline models, trained on the ``simple" formulation, grouped by each disaster contained within the test set. Table \ref{tab:model_perf_disaster_full} shows the performance of the baseline models, trained on the ``full" formulation, grouped by each disaster contained within the test set. To measure the performance of the model on each label Table \ref{tab:model_perf_label_simple} and Table \ref{tab:model_perf_label_full} are provided. Table \ref{tab:model_perf_label_simple} provides the performance of the model on each label within the ``simple" formulation. Table \ref{tab:model_perf_label_full} provides the performance of the model on each label within the ``full" formulation.

\begin{table*}[]
\resizebox{\textwidth}{!}{
\begin{tabular}{|ll|ll|ll|ll|ll|ll|}
\hline
\textbf{}                                                                      & \textbf{}          & \multicolumn{2}{c|}{\textbf{Global}} & \multicolumn{2}{c|}{\textbf{\begin{tabular}[c]{@{}c@{}}Hurricane\\ Idalia\end{tabular}}} & \multicolumn{2}{c|}{\textbf{\begin{tabular}[c]{@{}c@{}}Mayfield\\ Tornado\end{tabular}}} & \multicolumn{2}{c|}{\textbf{\begin{tabular}[c]{@{}c@{}}Hurricane\\ Michael\end{tabular}}} & \multicolumn{2}{c|}{\textbf{\begin{tabular}[c]{@{}c@{}}Mussett \\ Bayou Fire\end{tabular}}} \\ \hline
\multicolumn{1}{|l|}{\textbf{Model}}                                           & \textbf{Alignment} & \textbf{IoU$_{km}$}    & \textbf{F1$_{km}$}    & \textbf{IoU$_{km}$}                              & \textbf{F1$_{km}$}                              & \textbf{IoU$_{km}$}                              & \textbf{F1$_{km}$}                              & \textbf{IoU$_{km}$}                               & \textbf{F1$_{km}$}                              & \textbf{IoU$_{km}$}                                & \textbf{F1$_{km}$}                               \\ \hline
\multicolumn{1}{|l|}{\multirow{2}{*}{UNet w/ Attention}}                       & Unadjusted         & 0.312             & 0.368            & 0.238                                       & 0.318                                      & 0.324                                       & 0.347                                      & 0.293                                        & 0.369                                      & 0.348                                         & 0.387                                       \\
\multicolumn{1}{|l|}{}                                                         & Adjusted           & 0.331             & 0.393            & 0.254                                       & 0.340                                      & 0.351                                       & 0.385                                      & 0.313                                        & 0.390                                      & 0.390                                         & 0.444                                       \\ \hline
\multicolumn{1}{|l|}{\multirow{2}{*}{UNet w/out Attention}}                    & Unadjusted         & 0.279             & 0.342            & 0.273                                       & 0.364                                      & 0.287                                       & 0.333                                      & 0.255                                        & 0.331                                      & 0.268                                         & 0.317                                       \\
\multicolumn{1}{|l|}{}                                                         & Adjusted           & 0.307             & 0.376            & 0.325                                       & 0.435                                      & 0.310                                       & 0.375                                      & 0.277                                        & 0.351                                      & 0.295                                         & 0.343                                       \\ \hline
\multicolumn{1}{|l|}{\multirow{2}{*}{ResNet101 + DeepLabv3Plus}}               & Unadjusted         & 0.143             & 0.234            & 0.055                                       & 0.100                                      & 0.141                                       & 0.206                                      & 0.192                                        & 0.313                                      & 0.118                                         & 0.183                                       \\
\multicolumn{1}{|l|}{}                                                         & Adjusted           & 0.165             & 0.264            & 0.071                                       & 0.127                                      & 0.158                                       & 0.230                                      & 0.222                                        & 0.352                                      & 0.154                                         & 0.226                                       \\ \hline
\multicolumn{1}{|l|}{\multirow{2}{*}{ResNet101 + PSPNet}}                      & Unadjusted         & 0.269             & 0.370            & 0.216                                       & 0.321                                      & 0.245                                       & 0.296                                      & 0.293                                        & 0.422                                      & 0.181                                         & 0.246                                       \\
\multicolumn{1}{|l|}{}                                                         & Adjusted           & 0.283             & 0.388            & 0.226                                       & 0.337                                      & 0.259                                       & 0.315                                      & 0.322                                        & 0.453                                      & 0.178                                         & 0.244                                       \\ \hline
\multicolumn{1}{|l|}{\multirow{2}{*}{ScaleMAE's ViT + Segmenter}}              & Unadjusted         & 0.033             & 0.064            & 0.040                                       & 0.072                                      & 0.007                                       & 0.014                                      & 0.073                                        & 0.130                                      & 0.010                                         & 0.019                                       \\
\multicolumn{1}{|l|}{}                                                         & Adjusted           & 0.039             & 0.074            & 0.051                                       & 0.051                                      & 0.010                                       & 0.019                                      & 0.073                                        & 0.132                                      & 0.011                                         & 0.021                                       \\ \hline
\multicolumn{1}{|l|}{\multirow{2}{*}{ScaleMAE's ViT (Pretrained) + Segmenter}} & Unadjusted         & 0.013             & 0.025            & 0.023                                       & 0.044                                      & 0.002                                       & 0.005                                      & 0.029                                        & 0.053                                      & 0.005                                         & 0.010                                       \\
\multicolumn{1}{|l|}{}                                                         & Adjusted           & 0.015             & 0.028            & 0.029                                       & 0.053                                      & 0.003                                       & 0.006                                      & 0.029                                        & 0.054                                      & 0.005                                         & 0.010                                       \\ \hline
\multicolumn{1}{|l|}{\multirow{2}{*}{ScaleMAE's ViT + UperNet}}                & Unadjusted         & 0.200             & 0.296            & 0.141                                       & 0.222                                      & 0.182                                       & 0.242                                      & 0.228                                        & 0.222                                      & 0.182                                         & 0.222                                       \\
\multicolumn{1}{|l|}{}                                                         & Adjusted           & 0.211             & 0.309            & 0.157                                       & 0.245                                      & 0.189                                       & 0.249                                      & 0.249                                        & 0.373                                      & 0.139                                         & 0.206                                       \\ \hline
\multicolumn{1}{|l|}{\multirow{2}{*}{ScaleMAE's ViT (Pretrained) + UperNet}}   & Unadjusted         & 0.185             & 0.284            & 0.118                                       & 0.195                                      & 0.142                                       & 0.205                                      & 0.273                                        & 0.399                                      & 0.130                                         & 0.193                                       \\
\multicolumn{1}{|l|}{}                                                         & Adjusted           & 0.209             & 0.313            & 0.143                                       & 0.240                                      & 0.163                                       & 0.228                                      & 0.317                                        & 0.446                                      & 0.139                                         & 0.204                                       \\ \hline
\multicolumn{1}{|l|}{\multirow{2}{*}{Random Baseline}}                         & Unadjusted         & 0.135             & 0.215            & 0.138                                       & 0.221                                      & 0.118                                       & 0.181                                      & 0.164                                        & 0.269                                      & 0.116                                         & 0.177                                       \\
\multicolumn{1}{|l|}{}                                                         & Adjusted           & 0.135             & 0.215            & 0.138                                       & 0.221                                      & 0.118                                       & 0.181                                      & 0.164                                        & 0.269                                      & 0.116                                         & 0.177                                       \\ \hline
\end{tabular}}
\caption{Model Performance for Baseline Models grouped by disaster in the ``Simple" configuration.}
\label{tab:model_perf_disaster_simple}
\end{table*}

\begin{table*}[]
\resizebox{\textwidth}{!}{
\begin{tabular}{|ll|ll|ll|ll|ll|ll|}
\hline
\textbf{}                                                                      & \textbf{}          & \multicolumn{2}{c|}{\textbf{Global}} & \multicolumn{2}{c|}{\textbf{\begin{tabular}[c]{@{}c@{}}Hurricane\\ Idalia\end{tabular}}} & \multicolumn{2}{c|}{\textbf{\begin{tabular}[c]{@{}c@{}}Mayfield\\ Tornado\end{tabular}}} & \multicolumn{2}{c|}{\textbf{\begin{tabular}[c]{@{}c@{}}Hurricane\\ Michael\end{tabular}}} & \multicolumn{2}{c|}{\textbf{\begin{tabular}[c]{@{}c@{}}Mussett \\ Bayou Fire\end{tabular}}} \\ \hline
\multicolumn{1}{|l|}{\textbf{Model}}                                           & \textbf{Alignment} & \textbf{IoU$_{km}$}    & \textbf{F1$_{km}$}    & \textbf{IoU$_{km}$}                              & \textbf{F1$_{km}$}                              & \textbf{IoU$_{km}$}                              & \textbf{F1$_{km}$}                              & \textbf{IoU$_{km}$}                               & \textbf{F1$_{km}$}                              & \textbf{IoU$_{km}$}                                & \textbf{F1$_{km}$}                               \\ \hline
\multicolumn{1}{|l|}{\multirow{2}{*}{UNet w/ Attention}}                       & Unadjusted         & 0.092             & 0.096            & 0.093                                       & 0.096                                      & 0.098                                       & 0.099                                      & 0.077                                        & 0.087                                      & 0.099                                         & 0.099                                       \\
\multicolumn{1}{|l|}{}                                                         & Adjusted           & 0.091             & 0.095            & 0.089                                       & 0.094                                      & 0.098                                       & 0.099                                      & 0.077                                        & 0.087                                      & 0.098                                         & 0.099                                       \\ \hline
\multicolumn{1}{|l|}{\multirow{2}{*}{UNet w/out Attention}}                    & Unadjusted         & 0.092             & 0.096            & 0.093                                       & 0.096                                      & 0.098                                       & 0.099                                      & 0.077                                        & 0.087                                      & 0.099                                         & 0.099                                       \\
\multicolumn{1}{|l|}{}                                                         & Adjusted           & 0.091             & 0.095            & 0.089                                       & 0.094                                      & 0.098                                       & 0.099                                      & 0.077                                        & 0.087                                      & 0.098                                         & 0.099                                       \\ \hline
\multicolumn{1}{|l|}{\multirow{2}{*}{ResNet101 + DeepLabv3Plus}}               & Unadjusted         & 0.076             & 0.096            & 0.081                                       & 0.103                                      & 0.074                                       & 0.086                                      & 0.067                                        & 0.091                                      & 0.061                                         & 0.076                                       \\
\multicolumn{1}{|l|}{}                                                         & Adjusted           & 0.081             & 0.103            & 0.085                                       & 0.108                                      & 0.075                                       & 0.087                                      & 0.076                                        & 0.099                                      & 0.060                                         & 0.075                                       \\ \hline
\multicolumn{1}{|l|}{\multirow{2}{*}{ResNet101 + PSPNet}}                      & Unadjusted         & 0.083             & 0.095            & 0.087                                       & 0.096                                      & 0.086                                       & 0.093                                      & 0.073                                        & 0.091                                      & 0.078                                         & 0.088                                       \\
\multicolumn{1}{|l|}{}                                                         & Adjusted           & 0.084             & 0.095            & 0.086                                       & 0.096                                      & 0.086                                       & 0.092                                      & 0.079                                        & 0.097                                      & 0.079                                         & 0.088                                       \\ \hline
\multicolumn{1}{|l|}{\multirow{2}{*}{ScaleMAE's ViT + Segmenter}}              & Unadjusted         & 0.092             & 0.096            & 0.093                                       & 0.096                                      & 0.098                                       & 0.099                                      & 0.077                                        & 0.087                                      & 0.099                                         & 0.099                                       \\
\multicolumn{1}{|l|}{}                                                         & Adjusted           & 0.091             & 0.095            & 0.089                                       & 0.094                                      & 0.098                                       & 0.099                                      & 0.077                                        & 0.087                                      & 0.098                                         & 0.099                                       \\ \hline
\multicolumn{1}{|l|}{\multirow{2}{*}{ScaleMAE's ViT (Pretrained) + Segmenter}} & Unadjusted         & 0.092             & 0.096            & 0.093                                       & 0.096                                      & 0.098                                       & 0.099                                      & 0.077                                        & 0.087                                      & 0.099                                         & 0.099                                       \\
\multicolumn{1}{|l|}{}                                                         & Adjusted           & 0.091             & 0.095            & 0.089                                       & 0.094                                      & 0.098                                       & 0.099                                      & 0.077                                        & 0.087                                      & 0.098                                         & 0.099                                       \\ \hline
\multicolumn{1}{|l|}{\multirow{2}{*}{ScaleMAE's ViT + UperNet}}                & Unadjusted         & 0.085             & 0.095            & 0.088                                       & 0.106                                      & 0.089                                       & 0.094                                      & 0.075                                        & 0.088                                      & 0.085                                         & 0.092                                       \\
\multicolumn{1}{|l|}{}                                                         & Adjusted           & 0.087             & 0.099            & 0.082                                       & 0.098                                      & 0.089                                       & 0.094                                      & 0.075                                        & 0.087                                      & 0.082                                         & 0.090                                       \\ \hline
\multicolumn{1}{|l|}{\multirow{2}{*}{ScaleMAE's ViT (Pretrained) + UperNet}}   & Unadjusted         & 0.085             & 0.094            & 0.090                                       & 0.102                                      & 0.089                                       & 0.094                                      & 0.073                                        & 0.086                                      & 0.082                                         & 0.090                                       \\
\multicolumn{1}{|l|}{}                                                         & Adjusted           & 0.087             & 0.098            & 0.094                                       & 0.110                                      & 0.090                                       & 0.095                                      & 0.074                                        & 0.087                                      & 0.083                                         & 0.091                                       \\ \hline
\multicolumn{1}{|l|}{\multirow{2}{*}{Random Baseline}}                         & Unadjusted         & 0.016             & 0.031            & 0.015                                       & 0.029                                      & 0.011                                       & 0.021                                      & 0.026                                        & 0.048                                      & 0.011                                         & 0.021                                       \\
\multicolumn{1}{|l|}{}                                                         & Adjusted           & 0.016             & 0.031            & 0.015                                       & 0.029                                      & 0.011                                       & 0.021                                      & 0.026                                        & 0.048                                      & 0.011                                         & 0.021                                       \\ \hline
\end{tabular}}
\caption{Model Performance for Baseline Models grouped by disaster in the ``Full" configuration.}
\label{tab:model_perf_disaster_full}
\end{table*}

\begin{table*}[]
\resizebox{\textwidth}{!}{
\begin{tabular}{|ll|ll|ll|ll|ll|}
\hline
\textbf{}                                                                      & \textbf{}          & \multicolumn{2}{c|}{\textbf{Global}} & \multicolumn{2}{c|}{\textbf{Partial}} & \multicolumn{2}{c|}{\textbf{Road Line}} & \multicolumn{2}{c|}{\textbf{Total}} \\ \hline
\multicolumn{1}{|l|}{\textbf{Model}}                                           & \textbf{Alignment} & \textbf{IoU$_{km}$}    & \textbf{F1$_{km}$}    & \textbf{IoU$_{km}$}     & \textbf{F1$_{km}$}    & \textbf{IoU$_{km}$}      & \textbf{F1$_{km}$}     & \textbf{IoU$_{km}$}    & \textbf{F1$_{km}$}   \\ \hline
\multicolumn{1}{|l|}{\multirow{2}{*}{UNet w/ Attention}}                       & Unadjusted         & 0.312             & 0.368            & 0.005              & 0.009            & 0.825               & 0.904             & 0.105             & 0.191           \\
\multicolumn{1}{|l|}{}                                                         & Adjusted           & 0.331             & 0.393            & 0.002              & 0.005            & 0.839               & 0.912             & 0.160             & 0.263           \\ \hline
\multicolumn{1}{|l|}{\multirow{2}{*}{UNet w/out Attention}}                    & Unadjusted         & 0.279             & 0.343            & 0.015              & 0.030            & 0.742               & 0.852             & 0.080             & 0.147           \\
\multicolumn{1}{|l|}{}                                                         & Adjusted           & 0.307             & 0.376            & 0.011              & 0.022            & 0.787               & 0.881             & 0.096             & 0.175           \\ \hline
\multicolumn{1}{|l|}{\multirow{2}{*}{ResNet101 + DeepLabv3Plus}}               & Unadjusted         & 0.143             & 0.234            & 0.048              & 0.092            & 0.306               & 0.469             & 0.075             & 0.140           \\
\multicolumn{1}{|l|}{}                                                         & Adjusted           & 0.165             & 0.264            & 0.061              & 0.114            & 0.338               & 0.506             & 0.095             & 0.173           \\ \hline
\multicolumn{1}{|l|}{\multirow{2}{*}{ResNet101 + PSPNet}}                      & Unadjusted         & 0.269             & 0.370            & 0.125              & 0.197            & 0.632               & 0.765             & 0.079             & 0.147           \\
\multicolumn{1}{|l|}{}                                                         & Adjusted           & 0.283             & 0.388            & 0.120              & 0.214            & 0.634               & 0.776             & 0.095             & 0.173           \\ \hline
\multicolumn{1}{|l|}{\multirow{2}{*}{ScaleMAE's ViT + Segmenter}}              & Unadjusted         & 0.033             & 0.064            & 0.044              & 0.084            & 0.000               & 0.000             & 0.047             & 0.106           \\
\multicolumn{1}{|l|}{}                                                         & Adjusted           & 0.039             & 0.074            & 0.046              & 0.088            & 0.000               & 0.000             & 0.071             & 0.133           \\ \hline
\multicolumn{1}{|l|}{\multirow{2}{*}{ScaleMAE's ViT (Pretrained) + Segmenter}} & Unadjusted         & 0.013             & 0.025            & 0.000              & 0.000            & 8x10\textsuperscript{-6}              & 2x10\textsuperscript{-6}            & 0.041             & 0.075           \\
\multicolumn{1}{|l|}{}                                                         & Adjusted           & 0.015             & 0.028            & 0.000              & 0.000            & 1x10\textsuperscript{-5}              & 2x10\textsuperscript{-5}            & 0.044             & 0.085           \\ \hline
\multicolumn{1}{|l|}{\multirow{2}{*}{ScaleMAE's ViT + UperNet}}                & Unadjusted         & 0.200             & 0.296            & 0.080              & 0.148            & 0.464               & 0.633             & 0.057             & 0.108           \\
\multicolumn{1}{|l|}{}                                                         & Adjusted           & 0.211             & 0.309            & 0.076              & 0.165            & 0.483               & 0.667             & 0.073             & 0.125           \\ \hline
\multicolumn{1}{|l|}{\multirow{2}{*}{ScaleMAE's ViT (Pretrained) + UperNet}}   & Unadjusted         & 0.185             & 0.284            & 0.095              & 0.174            & 0.407               & 0.578             & 0.052             & 0.099           \\
\multicolumn{1}{|l|}{}                                                         & Adjusted           & 0.209             & 0.313            & 0.105              & 0.190            & 0.454               & 0.625             & 0.067             & 0.125           \\ \hline
\multicolumn{1}{|l|}{\multirow{2}{*}{Random Baseline}}                         & Unadjusted         & 0.135             & 0.215            & 0.044              & 0.107            & 0.324               & 0.484             & 0.038             & 0.074           \\
\multicolumn{1}{|l|}{}                                                         & Adjusted           & 0.135             & 0.215            & 0.044              & 0.107            & 0.324               & 0.484             & 0.038             & 0.074           \\ \hline
\end{tabular}}
\caption{Model Performance for Baseline Models for each label in the ``Simple" configuration.}
\label{tab:model_perf_label_simple}
\end{table*}


\begin{table*}[]
\resizebox{\textwidth}{!}{
\begin{tabular}{|ll|rr|rr|rr|rr|rr|rr}
\hline
\textbf{}                                                                      & \textbf{}          & \multicolumn{2}{c|}{\textbf{\begin{tabular}[c]{@{}c@{}}Macro\\ Average\end{tabular}}}     & \multicolumn{2}{c|}{\textbf{\begin{tabular}[c]{@{}c@{}}Not Able \\ To Determine\end{tabular}}} & \multicolumn{2}{c|}{\textbf{\begin{tabular}[c]{@{}c@{}}Partial\\ Flooding\end{tabular}}}  & \multicolumn{2}{c|}{\textbf{\begin{tabular}[c]{@{}c@{}}Partial\\ Obstruction\end{tabular}}} & \multicolumn{2}{c|}{\textbf{\begin{tabular}[c]{@{}c@{}}Partial Road \\ Condition\end{tabular}}} & \multicolumn{2}{c|}{\textbf{\begin{tabular}[c]{@{}c@{}}Partial\\ Particulate\end{tabular}}} \\ \hline
\multicolumn{1}{|l|}{\textbf{Model}}                                           & \textbf{Alignment} & \multicolumn{1}{l}{\textbf{IoU\textsubscript{km}}}          & \multicolumn{1}{l|}{\textbf{F1\textsubscript{km}}}          & \multicolumn{1}{l}{\textbf{IoU\textsubscript{km}}}             & \multicolumn{1}{l|}{\textbf{F1\textsubscript{km}}}            & \multicolumn{1}{l}{\textbf{IoU\textsubscript{km}}}           & \multicolumn{1}{l|}{\textbf{F1\textsubscript{km}}}         & \multicolumn{1}{l}{\textbf{IoU\textsubscript{km}}}           & \multicolumn{1}{l|}{\textbf{F1\textsubscript{km}}}           & \multicolumn{1}{l}{\textbf{IoU\textsubscript{km}}}             & \multicolumn{1}{l|}{\textbf{F1\textsubscript{km}}}             & \multicolumn{1}{l}{\textbf{IoU\textsubscript{km}}}           & \multicolumn{1}{l|}{\textbf{F1\textsubscript{km}}}           \\ \hline
\multicolumn{1}{|l|}{\multirow{2}{*}{UNet w/ Attention}}                       & Unadjusted         & 0.092                                       & 0.096                                       & 0.000                                          & 0.000                                         & 0.000                                        & 0.000                                      & 0.000                                        & 0.000                                        & 0.000                                          & 0.000                                          & 0.000                                        & \multicolumn{1}{r|}{0.000}                   \\
\multicolumn{1}{|l|}{}                                                         & Adjusted           & 0.091                                       & 0.095                                       & 0.000                                          & 0.000                                         & 0.000                                        & 0.000                                      & 0.000                                        & 0.000                                        & 0.000                                          & 0.000                                          & 0.000                                        & \multicolumn{1}{r|}{0.000}                   \\ \hline
\multicolumn{1}{|l|}{\multirow{2}{*}{UNet w/out Attention}}                    & Unadjusted         & 0.092                                       & 0.096                                       & 0.000                                          & 0.000                                         & 0.000                                        & 0.000                                      & 0.000                                        & 0.000                                        & 0.000                                          & 0.000                                          & 0.000                                        & \multicolumn{1}{r|}{0.000}                   \\
\multicolumn{1}{|l|}{}                                                         & Adjusted           & 0.091                                       & 0.095                                       & 0.000                                          & 0.000                                         & 0.000                                        & 0.000                                      & 0.000                                        & 0.000                                        & 0.000                                          & 0.000                                          & 0.000                                        & \multicolumn{1}{r|}{0.000}                   \\ \hline
\multicolumn{1}{|l|}{\multirow{2}{*}{ResNet101 + DeepLabv3Plus}}               & Unadjusted         & 0.076                                       & 0.096                                       & 0.000                                          & 0.000                                         & 0.000                                        & 0.000                                      & 0.000                                        & 0.000                                        & 0.000                                          & 0.000                                          & 0.000                                        & \multicolumn{1}{r|}{0.000}                   \\
\multicolumn{1}{|l|}{}                                                         & Adjusted           & 0.081                                       & 0.103                                       & 0.000                                          & 0.000                                         & 0.000                                        & 0.000                                      & 0.000                                        & 0.000                                        & 0.000                                          & 0.000                                          & 0.000                                        & \multicolumn{1}{r|}{0.000}                   \\ \hline
\multicolumn{1}{|l|}{\multirow{2}{*}{ResNet101 + PSPNet}}                      & Unadjusted         & 0.083                                       & 0.095                                       & 0.000                                          & 0.000                                         & 0.000                                        & 0.000                                      & 0.000                                        & 0.000                                        & 0.000                                          & 0.000                                          & 0.000                                        & \multicolumn{1}{r|}{0.000}                   \\
\multicolumn{1}{|l|}{}                                                         & Adjusted           & 0.084                                       & 0.095                                       & 0.000                                          & 0.000                                         & 0.000                                        & 0.000                                      & 0.000                                        & 0.000                                        & 0.000                                          & 0.000                                          & 0.000                                        & \multicolumn{1}{r|}{0.000}                   \\ \hline
\multicolumn{1}{|l|}{\multirow{2}{*}{ScaleMAE's ViT + Segmenter}}              & Unadjusted         & 0.092                                       & 0.096                                       & 0.000                                          & 0.000                                         & 0.000                                        & 0.000                                      & 0.000                                        & 0.000                                        & 0.000                                          & 0.000                                          & 0.000                                        & \multicolumn{1}{r|}{0.000}                   \\
\multicolumn{1}{|l|}{}                                                         & Adjusted           & 0.091                                       & 0.095                                       & 0.000                                          & 0.000                                         & 0.000                                        & 0.000                                      & 0.000                                        & 0.000                                        & 0.000                                          & 0.000                                          & 0.000                                        & \multicolumn{1}{r|}{0.000}                   \\ \hline
\multicolumn{1}{|l|}{\multirow{2}{*}{ScaleMAE's ViT (Pretrained) + Segmenter}} & Unadjusted         & 0.092                                       & 0.096                                       & 0.000                                          & 0.000                                         & 0.000                                        & 0.000                                      & 0.000                                        & 0.000                                        & 0.000                                          & 0.000                                          & 0.000                                        & \multicolumn{1}{r|}{0.000}                   \\
\multicolumn{1}{|l|}{}                                                         & Adjusted           & 0.091                                       & 0.095                                       & 0.000                                          & 0.000                                         & 0.000                                        & 0.000                                      & 0.000                                        & 0.000                                        & 0.000                                          & 0.000                                          & 0.000                                        & \multicolumn{1}{r|}{0.000}                   \\ \hline
\multicolumn{1}{|l|}{\multirow{2}{*}{ScaleMAE's ViT + UperNet}}                & Unadjusted         & 0.085                                       & 0.095                                       & 0.000                                          & 0.000                                         & 0.000                                        & 0.000                                      & 0.000                                        & 0.000                                        & 0.000                                          & 0.000                                          & 0.000                                        & \multicolumn{1}{r|}{0.000}                   \\
\multicolumn{1}{|l|}{}                                                         & Adjusted           & 0.087                                       & 0.099                                       & 0.000                                          & 0.000                                         & 0.000                                        & 0.000                                      & 0.000                                        & 0.000                                        & 0.000                                          & 0.000                                          & 0.000                                        & \multicolumn{1}{r|}{0.000}                   \\ \hline
\multicolumn{1}{|l|}{\multirow{2}{*}{ScaleMAE's ViT (Pretrained) + UperNet}}   & Unadjusted         & 0.085                                       & 0.094                                       & 0.000                                          & 0.000                                         & 0.000                                        & 0.000                                      & 0.000                                        & 0.000                                        & 0.000                                          & 0.000                                          & 0.000                                        & \multicolumn{1}{r|}{0.000}                   \\
\multicolumn{1}{|l|}{}                                                         & Adjusted           & 0.087                                       & 0.098                                       & 0.000                                          & 0.000                                         & 0.000                                        & 0.000                                      & 0.000                                        & 0.000                                        & 0.000                                          & 0.000                                          & 0.000                                        & \multicolumn{1}{r|}{0.000}                   \\ \hline
\multicolumn{1}{|l|}{\multirow{2}{*}{Random Baseline}}                         & Unadjusted         & 0.016                                       & 0.031                                       & 0.005                                          & 0.010                                         & 0.001                                        & 0.001                                      & 0.009                                        & 0.017                                        & 0.006                                          & 0.012                                          & 0.012                                        & \multicolumn{1}{r|}{0.039}                   \\
\multicolumn{1}{|l|}{}                                                         & Adjusted           & 0.016                                       & 0.031                                       & 0.005                                          & 0.010                                         & 0.001                                        & 0.001                                      & 0.009                                        & 0.017                                        & 0.006                                          & 0.012                                          & 0.012                                        & \multicolumn{1}{r|}{0.039}                   \\ \hline
\multicolumn{2}{|l|}{\multirow{2}{*}{}}                                                             & \multicolumn{2}{c|}{\textbf{\begin{tabular}[c]{@{}c@{}}Total\\ Particulate\end{tabular}}} & \multicolumn{2}{c|}{\textbf{Road Line}}                                                        & \multicolumn{2}{c|}{\textbf{\begin{tabular}[c]{@{}c@{}}Total\\ Destruction\end{tabular}}} & \multicolumn{2}{c|}{\textbf{\begin{tabular}[c]{@{}c@{}}Total\\ Flooding\end{tabular}}}      & \multicolumn{2}{c|}{\textbf{\begin{tabular}[c]{@{}c@{}}Total\\ Obstruction\end{tabular}}}       & \multicolumn{2}{c}{\textbf{}}                                                               \\ \cline{3-12}
\multicolumn{2}{|l|}{}                                                                              & \multicolumn{1}{l}{\textbf{IoU\textsubscript{km}}}          & \multicolumn{1}{l|}{\textbf{F1\textsubscript{km}}}          & \multicolumn{1}{l}{\textbf{IoU\textsubscript{km}}}             & \multicolumn{1}{l|}{\textbf{F1\textsubscript{km}}}            & \multicolumn{1}{l}{\textbf{IoU\textsubscript{km}}}           & \multicolumn{1}{l|}{\textbf{F1\textsubscript{km}}}         & \multicolumn{1}{l}{\textbf{IoU\textsubscript{km}}}           & \multicolumn{1}{l|}{\textbf{F1\textsubscript{km}}}           & \multicolumn{1}{l}{\textbf{IoU\textsubscript{km}}}             & \multicolumn{1}{l|}{\textbf{F1\textsubscript{km}}}             & \multicolumn{1}{l}{\textbf{}}                & \multicolumn{1}{l}{\textbf{}}                \\ \cline{1-12}
\multicolumn{1}{|l|}{\multirow{2}{*}{UNet w/ Attention}}                       & Unadjusted         & 0.000                                       & 0.000                                       & 0.923                                          & 0.960                                         & 0.000                                        & 0.000                                      & 0.000                                        & 0.000                                        & 0.000                                          & 0.000                                          &                                              &                                              \\
\multicolumn{1}{|l|}{}                                                         & Adjusted           & 0.000                                       & 0.000                                       & 0.910                                          & 0.953                                         & 0.000                                        & 0.000                                      & 0.000                                        & 0.000                                        & 0.000                                          & 0.000                                          &                                              &                                              \\ \cline{1-12}
\multicolumn{1}{|l|}{\multirow{2}{*}{UNet w/out Attention}}                    & Unadjusted         & 0.000                                       & 0.000                                       & \multicolumn{1}{r|}{0.923}                     & 0.960                                         & 0.000                                        & 0.000                                      & 0.000                                        & 0.000                                        & 0.000                                          & 0.000                                          &                                              &                                              \\ \cline{5-5}
\multicolumn{1}{|l|}{}                                                         & Adjusted           & 0.000                                       & 0.000                                       & 0.910                                          & 0.953                                         & 0.000                                        & 0.000                                      & 0.000                                        & 0.000                                        & 0.000                                          & 0.000                                          &                                              &                                              \\ \cline{1-12}
\multicolumn{1}{|l|}{\multirow{2}{*}{ResNet101 + DeepLabv3Plus}}               & Unadjusted         & 0.029                                       & 0.057                                       & 0.682                                          & 0.811                                         & 0.000                                        & 0.000                                      & 0.048                                        & 0.093                                        & 0.000                                          & 0.000                                          &                                              &                                              \\
\multicolumn{1}{|l|}{}                                                         & Adjusted           & 0.033                                       & 0.063                                       & 0.706                                          & 0.828                                         & 0.000                                        & 0.000                                      & 0.072                                        & 0.134                                        & 0.000                                          & 0.000                                          &                                              &                                              \\ \cline{1-12}
\multicolumn{1}{|l|}{\multirow{2}{*}{ResNet101 + PSPNet}}                      & Unadjusted         & 0.027                                       & 0.064                                       & 0.807                                          & 0.879                                         & 0.000                                        & 0.000                                      & 0.000                                        & 0.000                                        & 0.000                                          & 0.000                                          &                                              &                                              \\
\multicolumn{1}{|l|}{}                                                         & Adjusted           & 0.030                                       & 0.058                                       & 0.812                                          & 0.896                                         & 0.000                                        & 0.000                                      & 0.000                                        & 0.000                                        & 0.000                                          & 0.000                                          &                                              &                                              \\ \cline{1-12}
\multicolumn{1}{|l|}{\multirow{2}{*}{ScaleMAE's ViT + Segmenter}}              & Unadjusted         & 0.000                                       & 0.000                                       & 0.923                                          & 0.960                                         & 0.000                                        & 0.000                                      & 0.000                                        & 0.000                                        & 0.000                                          & 0.000                                          &                                              &                                              \\
\multicolumn{1}{|l|}{}                                                         & Adjusted           & 0.000                                       & 0.000                                       & 0.910                                          & 0.953                                         & 0.000                                        & 0.000                                      & 0.000                                        & 0.000                                        & 0.000                                          & 0.000                                          &                                              &                                              \\ \cline{1-12}
\multicolumn{1}{|l|}{\multirow{2}{*}{ScaleMAE's ViT (Pretrained) + Segmenter}} & Unadjusted         & 0.000                                       & 0.000                                       & 0.923                                          & 0.960                                         & 0.000                                        & 0.000                                      & 0.000                                        & 0.000                                        & 0.000                                          & 0.000                                          &                                              &                                              \\
\multicolumn{1}{|l|}{}                                                         & Adjusted           & 0.000                                       & 0.000                                       & 0.910                                          & 0.953                                         & 0.000                                        & 0.000                                      & 0.000                                        & 0.000                                        & 0.000                                          & 0.000                                          &                                              &                                              \\ \cline{1-12}
\multicolumn{1}{|l|}{\multirow{2}{*}{ScaleMAE's ViT + UperNet}}                & Unadjusted         & 0.000                                       & 0.000                                       & 0.822                                          & 0.902                                         & 0.000                                        & 0.000                                      & 0.025                                        & 0.049                                        & 0.000                                          & 0.000                                          &                                              &                                              \\
\multicolumn{1}{|l|}{}                                                         & Adjusted           & 0.000                                       & 0.000                                       & 0.821                                          & 0.905                                         & 0.000                                        & 0.000                                      & 0.017                                        & 0.088                                        & 0.000                                          & 0.000                                          &                                              &                                              \\ \cline{1-12}
\multicolumn{1}{|l|}{\multirow{2}{*}{ScaleMAE's ViT (Pretrained) + UperNet}}   & Unadjusted         & 0.000                                       & 0.000                                       & 0.822                                          & 0.912                                         & 0.000                                        & 0.000                                      & 0.007                                        & 0.032                                        & 0.000                                          & 0.000                                          &                                              &                                              \\
\multicolumn{1}{|l|}{}                                                         & Adjusted           & 7x10\textsuperscript{-5}                                      & 1x10\textsuperscript{-4}                                      & 0.840                                          & 0.913                                         & 0.012                                        & 0.064                                      & 0.000                                        & 0.000                                        & 0.000                                          & 0.000                                          &                                              &                                              \\ \cline{1-12}
\multicolumn{1}{|l|}{\multirow{2}{*}{Random Baseline}}                         & Unadjusted         & 0.014                                       & 0.027                                       & 0.100                                          & 0.179                                         & 0.001                                        & 0.001                                      & 0.002                                        & 0.005                                        & 0.009                                          & 0.018                                          &                                              &                                              \\
\multicolumn{1}{|l|}{}                                                         & Adjusted           & 0.014                                       & 0.027                                       & 0.100                                          & 0.179                                         & 0.001                                        & 0.001                                      & 0.002                                        & 0.005                                        & 0.009                                          & 0.018                                          &                                              &                                              \\ \cline{1-12}
\end{tabular}}
\caption{Model Performance for Baseline Models for each label in the ``Full" configuration.}
\label{tab:model_perf_label_full}
\end{table*}

\section*{Ethical Statement}

All imagery in this work has been affirmatively released by the agencies having jurisdiction; it was collected in accordance with the appropriate FAA regulations and guidance, and it was collected at the direction of agencies having jurisdiction. In conjunction with this release, the agencies having jurisdiction screened all imagery and withheld imagery that they did not want to be released publicly. 

All imagery considered in this work was collected at the direction of agencies having jurisdiction, and thus, it captures the operational distribution of data. As a result, this work was scoped such that commentary on the nature of data collection and commentary on how practitioners direct data collection, as it relates to potential over- or undersampling of specific geographies and socioeconomic statuses, is out of scope for this work. When deploying ML systems without human oversight, such analyses are crucial to ensure that models do not exacerbate biases that they have learned. Future work to measure and mitigate biases learned by models trained on this data and based on the operational distribution of data will be critical to ensure models can operate with minimal technical oversight in the future. At the time of writing, the specific ethical implications of false positives and false negatives are currently being explored in an effort to mitigate potential negative consequences created by these models and labels.

Readers are strongly discouraged from deploying these models in disaster operations without coordination with the authors of this work. As discussed in this work, these models have limitations and faults, and direct introduction of these models, without trained human oversight, risks the unmitigated introduction of model biases in operational environments. Should readers wish to utilize these models in practice, please contact the authors directly to discuss the deployment considerations in detail.

\section*{Acknowledgements}
This work is supported by the AI Research Institutes Program funded by the National Science Foundation under the AI Institute for Societal Decision Making (NSF AI-SDM), Award No. 2229881, and under ``Datasets for Uncrewed Aerial System (UAS) and Remote Responder Performance from Hurricane Ian" Award No. 2306453. The authors thank the Florida State Emergency Response Team, FL-UAS1 task force, and Florida State University for their support, and the Winchester Thurston School, Ball High School, Bryan Collegiate High School, and Rudder High School for their annotation efforts.

\bibliography{aaai2026}

@article{manzini2024crasar,
  title={CRASAR-U-DROIDs: A Large Scale Benchmark Dataset for Building Alignment and Damage Assessment in Georectified sUAS Imagery},
  author={Manzini, Thomas and Perali, Priyankari and Karnik, Raisa and Murphy, Robin},
  journal={arXiv preprint arXiv:2407.17673},
  year={2024}
}

@article{manzini2024non,
  title={Non-Uniform Spatial Alignment Errors in sUAS Imagery From Wide-Area Disasters},
  author={Manzini, Thomas and Perali, Priyankari and Karnik, Raisa and Godbole, Mihir and Abdullah, Hasnat and Murphy, Robin},
  journal={arXiv preprint arXiv:2405.06593},
  year={2024}
}

@article{rahnemoonfar2021floodnet,
  title={Floodnet: A high resolution aerial imagery dataset for post flood scene understanding},
  author={Rahnemoonfar, Maryam and Chowdhury, Tashnim and Sarkar, Argho and Varshney, Debvrat and Yari, Masoud and Murphy, Robin Roberson},
  journal={IEEE Access},
  volume={9},
  pages={89644--89654},
  year={2021},
  publisher={IEEE}
}

@article{takyi2025towards,
  title={Towards Sustainable and Resilient Infrastructure: Hurricane-Induced Roadway Closure and Accessibility Assessment in Florida Using Machine Learning},
  author={Takyi, Samuel and Antwi, Richard Boadu and Ozguven, Eren Erman and Okine, Leslie and Moses, Ren},
  journal={Sustainability},
  volume={17},
  number={9},
  pages={3909},
  year={2025},
  publisher={MDPI}
}

@article{rahnemoonfar2023rescuenet,
  title={RescueNet: a high resolution UAV semantic segmentation dataset for natural disaster damage assessment},
  author={Rahnemoonfar, Maryam and Chowdhury, Tashnim and Murphy, Robin},
  journal={Scientific data},
  volume={10},
  number={1},
  pages={913},
  year={2023},
  publisher={Nature Publishing Group UK London}
}

@misc{OpenStreetMap,
   author = {{OpenStreetMap contributors}},
   title = {{Planet dump retrieved from https://planet.osm.org }},
   howpublished = "\url{ https://www.openstreetmap.org } [Accessed: 2025-11-13]",
   year = {2024}
 }

@inproceedings{hansch2022spacenet,
  title={Spacenet 8-the detection of flooded roads and buildings},
  author={H{\"a}nsch, Ronny and Arndt, Jacob and Lunga, Dalton and Gibb, Matthew and Pedelose, Tyler and Boedihardjo, Arnold and Petrie, Desiree and Bacastow, Todd M},
  booktitle={Proceedings of the IEEE/CVF conference on computer vision and pattern recognition},
  pages={1472--1480},
  year={2022}
}

@misc{Labelbox2024,
  author = {Labelbox},
  title = {Labelbox},
  year = {2024},
  howpublished = "\url{ https://labelbox.com } [Accessed: 2025-11-13]"
}

@misc{Debby2024,
  author = {NOAA},
  title = {Debby resources: The latest storm forecasts, maps, imagery and more},
  year = {2024},
  howpublished = "\url{ https://www.noaa.gov/debby } [Accessed: 2025-11-13]"

}

@misc{Helene2024,
  author = {NOAA},
  title = {Helene resources: The latest storm forecasts, maps, imagery and more},
  year = {2024},
}

@article{maiti2022effect,
  title={Effect of label noise in semantic segmentation of high resolution aerial images and height data},
  author={Maiti, A and Oude Elberink, SJ and Vosselman, G},
  journal={ISPRS Annals of the Photogrammetry, Remote Sensing and Spatial Information Sciences},
  volume={2},
  pages={275--282},
  year={2022},
  publisher={Copernicus GmbH}
}

@article{vargas2019correcting,
  title={Correcting rural building annotations in OpenStreetMap using convolutional neural networks},
  author={Vargas-Mu{\~n}oz, John E and Lobry, Sylvain and Falc{\~a}o, Alexandre X and Tuia, Devis},
  journal={ISPRS journal of photogrammetry and remote sensing},
  volume={147},
  pages={283--293},
  year={2019},
  publisher={Elsevier}
}

@inproceedings{urabe2007analysis,
  title={Analysis of road blockage after disaster using aerial images},
  author={Urabe, Kazuya and Saji, Hitoshi},
  booktitle={SICE Annual Conference 2007},
  pages={1795--1798},
  year={2007},
  organization={IEEE}
}

@inproceedings{korkmaz2016path,
  title={Path planning for rescue vehicles via segmented satellite disaster images and GPS road map},
  author={Korkmaz, S Aytac and Poyraz, Mustafa},
  booktitle={2016 9th International Congress on Image and Signal Processing, BioMedical Engineering and Informatics (CISP-BMEI)},
  pages={145--150},
  year={2016},
  organization={IEEE}
}

@article{yang2020extraction,
  title={Extraction of road blockage information for the Jiuzhaigou earthquake based on a convolution neural network and very-high-resolution satellite images},
  author={Yang, Baolin and Wang, Shixin and Zhou, Yi and Wang, Futao and Hu, Qiao and Chang, Ying and Zhao, Qing},
  journal={Earth Science Informatics},
  volume={13},
  pages={115--127},
  year={2020},
  publisher={Springer}
}

@article{majidifard2020pavement,
  title={Pavement image datasets: A new benchmark dataset to classify and densify pavement distresses},
  author={Majidifard, Hamed and Jin, Peng and Adu-Gyamfi, Yaw and Buttlar, William G},
  journal={Transportation Research Record},
  volume={2674},
  number={2},
  pages={328--339},
  year={2020},
  publisher={SAGE Publications Sage CA: Los Angeles, CA}
}

@article{sabouri2023machine,
  title={Machine learning based readmission and mortality prediction in heart failure patients},
  author={Sabouri, Maziar and Rajabi, Ahmad Bitarafan and Hajianfar, Ghasem and Gharibi, Omid and Mohebi, Mobin and Avval, Atlas Haddadi and Naderi, Nasim and Shiri, Isaac},
  journal={Scientific Reports},
  volume={13},
  number={1},
  pages={18671},
  year={2023},
  publisher={Nature Publishing Group UK London}
}

@article{ren2024annotated,
  title={An annotated street view image dataset for automated road damage detection},
  author={Ren, Miao and Zhang, Xianfeng and Zhi, Xiaobo and Wei, Yuanjia and Feng, Ziyuan},
  journal={Scientific Data},
  volume={11},
  number={1},
  pages={407},
  year={2024},
  publisher={Nature Publishing Group UK London}
}

@article{yan2023uav,
  title={UAV-PDD2023: A benchmark dataset for pavement distress detection based on UAV images},
  author={Yan, Haohui and Zhang, Junfei},
  journal={Data in Brief},
  volume={51},
  pages={109692},
  year={2023},
  publisher={Elsevier}
}

@inproceedings{corsar2015transport,
  title={The transport disruption ontology},
  author={Corsar, David and Markovic, Milan and Edwards, Peter and Nelson, John D},
  booktitle={The Semantic Web-ISWC 2015: 14th International Semantic Web Conference, Bethlehem, PA, USA, October 11-15, 2015, Proceedings, Part II 14},
  pages={329--336},
  year={2015},
  organization={Springer}
}

@article{oktay2018attention,
  title={Attention u-net: Learning where to look for the pancreas},
  author={Oktay, Ozan and Schlemper, Jo and Folgoc, Loic Le and Lee, Matthew and Heinrich, Mattias and Misawa, Kazunari and Mori, Kensaku and McDonagh, Steven and Hammerla, Nils Y and Kainz, Bernhard and others},
  journal={arXiv preprint arXiv:1804.03999},
  year={2018}
}

@article{Pi2020, author = {Yalong Pi and Nipun D. Nath and Amir H. Behzadan}, title = {Convolutional neural networks for object detection in aerial imagery for disaster response and recovery}, journal = {Advanced Engineering Informatics}, volume = {43}, year = {2020}, pages = {101009}
}

@book{mitigation2007,
  title     = "Mitigation Strategies For Design Exceptions",
  author    = "Stein, William J. and Neuman, Timothy R.",
  year      = 2007,
  publisher = "US Department of Transportation",
  chapter   = "Chapter 3"
}

@inproceedings{hargis2024search,
  title={Search and rescue base of operation prioritization with aerial orthomosaics},
  author={Hargis, Alyssa and Rao, Ananya and Choset, Howie},
  booktitle={2024 IEEE International Symposium on Safety Security Rescue Robotics (SSRR)},
  pages={204--209},
  year={2024},
  organization={IEEE}
}

@inproceedings{alam2025harnessing,
  title={Harnessing Robotic Scouts for Resilient Evacuation Policies in Disaster Scenarios},
  author={Alam, Tauhidul and Quader, Sufi N and Islam, Sadman and Newaz, Abdullah Al Redwan},
  booktitle={2025 22nd International Conference on Ubiquitous Robots (UR)},
  pages={351--356},
  year={2025},
  organization={IEEE}
}

@inproceedings{manzini2023wireless,
  title={Wireless network demands of data products from small uncrewed aerial systems at Hurricane Ian},
  author={Manzini, Thomas and Murphy, Robin and Merrick, David and Adams, Justin},
  booktitle={2023 IEEE/RSJ International Conference on Intelligent Robots and Systems (IROS)},
  pages={9941--9946},
  year={2023},
  organization={IEEE}
}

@inproceedings{chen2018encoder,
  title={Encoder-decoder with atrous separable convolution for semantic image segmentation},
  author={Chen, Liang-Chieh and Zhu, Yukun and Papandreou, George and Schroff, Florian and Adam, Hartwig},
  booktitle={Proceedings of the European conference on computer vision (ECCV)},
  pages={801--818},
  year={2018}
}

@inproceedings{zhao2017pyramid,
  title={Pyramid scene parsing network},
  author={Zhao, Hengshuang and Shi, Jianping and Qi, Xiaojuan and Wang, Xiaogang and Jia, Jiaya},
  booktitle={Proceedings of the IEEE conference on computer vision and pattern recognition},
  pages={2881--2890},
  year={2017}
}

@inproceedings{xiao2018unified,
  title={Unified perceptual parsing for scene understanding},
  author={Xiao, Tete and Liu, Yingcheng and Zhou, Bolei and Jiang, Yuning and Sun, Jian},
  booktitle={Proceedings of the European conference on computer vision (ECCV)},
  pages={418--434},
  year={2018}
}

@inproceedings{strudel2021segmenter,
  title={Segmenter: Transformer for semantic segmentation},
  author={Strudel, Robin and Garcia, Ricardo and Laptev, Ivan and Schmid, Cordelia},
  booktitle={Proceedings of the IEEE/CVF international conference on computer vision},
  pages={7262--7272},
  year={2021}
}

@inproceedings{he2016deep,
  title={Deep residual learning for image recognition},
  author={He, Kaiming and Zhang, Xiangyu and Ren, Shaoqing and Sun, Jian},
  booktitle={Proceedings of the IEEE conference on computer vision and pattern recognition},
  pages={770--778},
  year={2016}
}

@inproceedings{reed2023scale,
  title={Scale-mae: A scale-aware masked autoencoder for multiscale geospatial representation learning},
  author={Reed, Colorado J and Gupta, Ritwik and Li, Shufan and Brockman, Sarah and Funk, Christopher and Clipp, Brian and Keutzer, Kurt and Candido, Salvatore and Uyttendaele, Matt and Darrell, Trevor},
  booktitle={Proceedings of the IEEE/CVF International Conference on Computer Vision},
  pages={4088--4099},
  year={2023}
}

@inproceedings{ronneberger2015u,
  title={U-net: Convolutional networks for biomedical image segmentation},
  author={Ronneberger, Olaf and Fischer, Philipp and Brox, Thomas},
  booktitle={International Conference on Medical image computing and computer-assisted intervention},
  pages={234--241},
  year={2015},
  organization={Springer}
}

@article{manzini2025challenges,
  title={Challenges and Research Directions from the Operational Use of a Machine Learning Damage Assessment System via Small Uncrewed Aerial Systems at Hurricanes Debby and Helene},
  author={Manzini, Thomas and Perali, Priyankari and Murphy, Robin R and Merrick, David},
  journal={arXiv preprint arXiv:2506.15890},
  year={2025}
}

@inproceedings{manzini2023quantitative,
  title={Quantitative data analysis: Crasar small unmanned aerial systems at hurricane ian},
  author={Manzini, Thomas and Murphy, Robin and Merrick, David},
  booktitle={2023 IEEE International Symposium on Safety, Security, and Rescue Robotics (SSRR)},
  pages={7--12},
  year={2023},
  organization={IEEE}
}

@inproceedings{jiang2024earthquakenet,
  title={EarthquakeNet: A High-Resolution UAV-Based Dataset for Earthquake Damage Assessment},
  author={Jiang, Shenlu and Bian, Yuxin and Wang, Yiran and Li, Xufeng and Liu, Zhankeng and Ren, Yi and Zhao, Yunxuan},
  booktitle={2024 IEEE International Conference on Image Processing (ICIP)},
  pages={55--61},
  year={2024},
  organization={IEEE}
}



\end{document}